\documentclass{article}
\usepackage{spconf,amsmath,graphicx}
\usepackage{amsfonts}
\usepackage{url}

\title{Principal Component Classification}
\name{Rozenn Dahyot \thanks{
This work was  funded by the SFI Research Centre ADAPT  (13/RC/2106 P2), and is co-funded by the European Regional Development Fund.}}
\address{Department of Computer Science, Maynooth University, Ireland}
\begin{document}
%
\maketitle
\begin{abstract}

We propose  to  directly compute classification estimates  
by learning  features encoded  with their class scores. 
Our resulting model
  has a  encoder-decoder structure suitable for  supervised learning, it is computationally efficient and performs well for classification  on several datasets.  
\end{abstract}
\section{Introduction}
\label{sec:intro}

The choice of data encoding  for defining inputs and outputs  of  machine learning pipelines 
contributes substantially to their performance. 
For instance, adding positional encoding in the inputs  have shown useful  for Convolutional Neural Networks  \cite{NEURIPS2018_60106888}  and for Neural radiance Fields   \cite{Nerf2021ACM}. 
Here, we propose to add  vectors of class scores as part of inputs to learn principal components suitable for predicting classification scores. 
Performance of our proposed frugal model is validated experimentally on datasets \textit{wine}, \textit{australian} \cite{Dua:UCI,10.5555/3045390.3045650} and \textit{MNIST} \cite{MNIST} for comparison  with 
 metric learning classification \cite{10.5555/1577069.1577078,10.5555/3045390.3045650,Eusipco2022Collas}, and deep learning \cite{MNIST,EfficientCapsNet2021}.

\section{Principal Component  Classification}

In supervised learning, we consider available a  dataset $\mathcal{B}=\lbrace(\mathbf{x}^{(i)},\mathbf{y}^{(i)})\rbrace_{i=1,\cdots,N}$ of $N$ observations
with $\mathbf{x}\in \mathbb{R}^{d_{\mathbf{x}}}$ denoting the  feature vector of dimension $d_{\mathbf{x}}$ and $\mathbf{y}\in \mathbb{R}^{n_c}$ the indicator class vector where $n_c$ is the number of classes. All coordinates of $\mathbf{y}^{(i)}$  are equal to zero at the exception of its coordinate $y_j^{(i)}$  
 that is equal to 1 if $\mathbf{y}^{(i)}$ is indicating that feature  vector $\mathbf{x}^{(i)}$ belongs to class  $j\in\lbrace 1,\cdots,n_c\rbrace$.
 
Principal Component Analysis (PCA) \cite{ClassBook2004} is a standard technique for dimension reduction 
often used in conjunction with classification techniques  \cite{10.5555/1577069.1577078}.
In PCA, the principal components correspond to the eigenvectors of the covariance matrix $\mathrm{\Sigma}$ ranked in descending order of  their associated  eigenvalues, where $\mathrm{\Sigma}=\frac{1}{N}\mathrm{X}\mathrm{X}^T$ and  $\mathrm{X}=\lbrack\mathbf{x}^{(1)}, \cdots,  \mathbf{x}^{(N)}\rbrack$. These principal components provide a orthonormal basis in the feature space. Retaining only the ones associated with the highest eigenvalues allow to project $\mathbf{x}$ in a very small dimensional eigenspace (data embedding). 
Such PCA based representation has been used for learning images of objects, to perform detection and registration \cite{MoghaddamPAMI1997,Dahyot_PAA,CVMP2013Arellano}, and has a probabilistic interpretation \cite{PPCA1999}.   
PCA for dimensionality reduction of the feature space ignores information from the class labels and we propose next a new data encoding suitable for  learning principal components that can be used for classification.

\begin{figure*}[t]
\begin{tabular}{|cccc|}
\hline
\includegraphics[width=.22\textwidth,trim={4.1cm 12.1cm 4.3cm 11.7cm},clip]{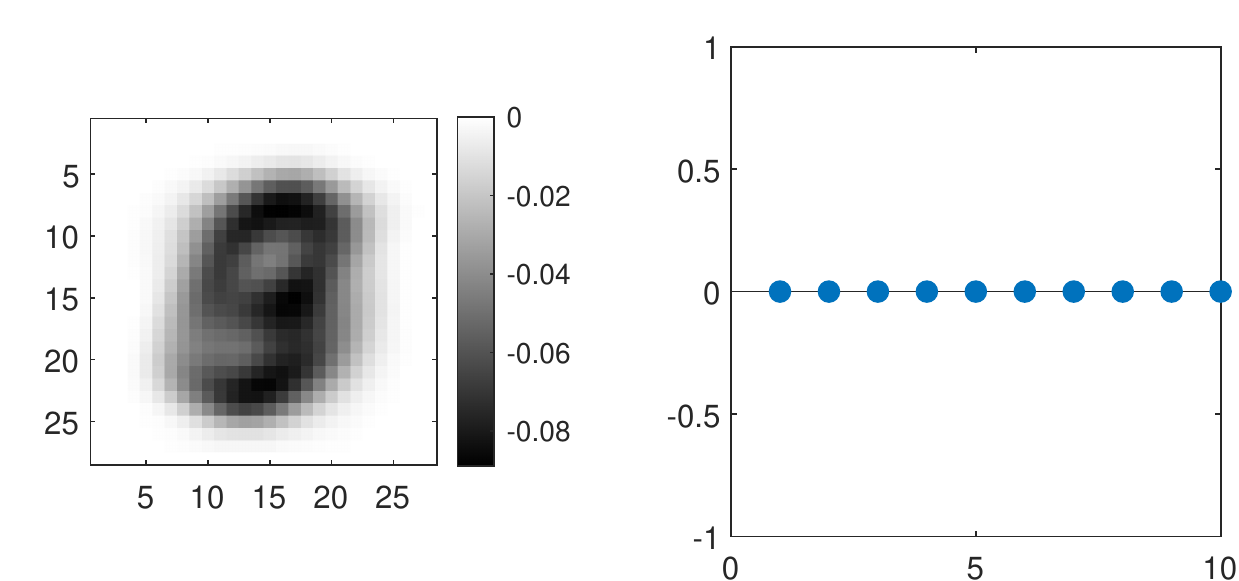}&
\includegraphics[width=.22\textwidth,trim={4.1cm 12.1cm 4.3cm 11.7cm},clip]{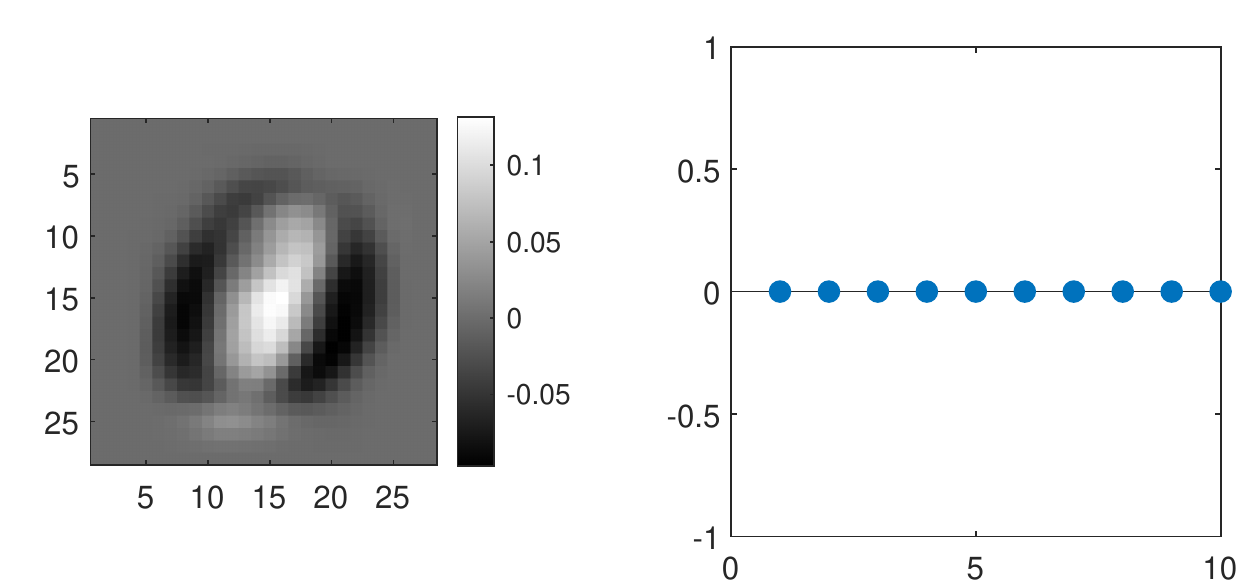}&
\includegraphics[width=.22\textwidth,trim={4.1cm 12.1cm 4.3cm 11.7cm},clip]{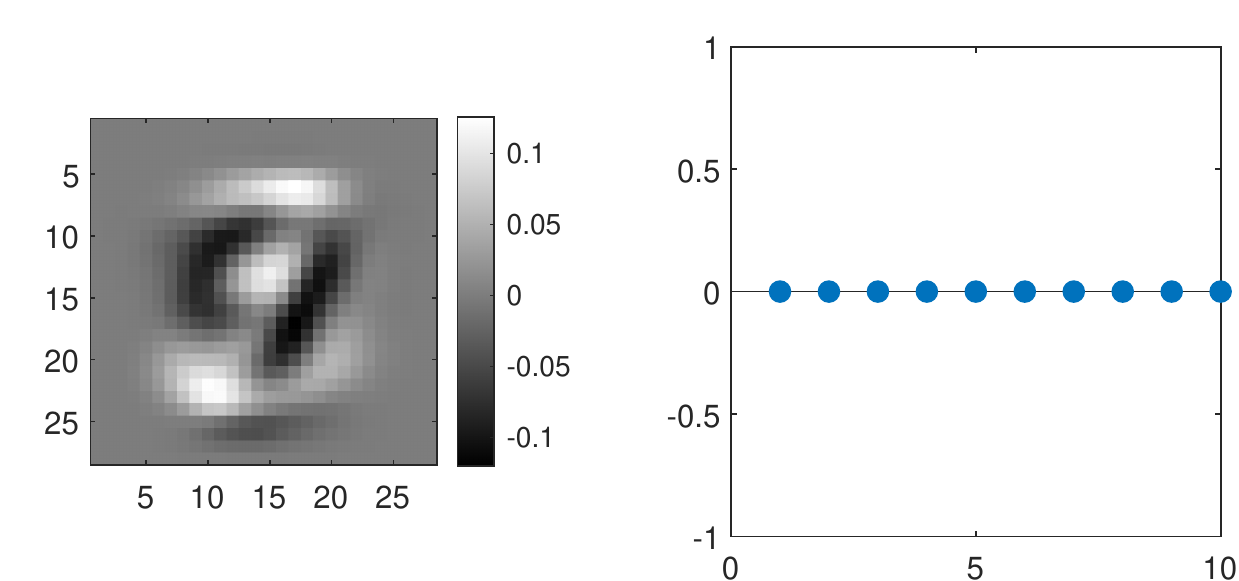}&
\includegraphics[width=.22\textwidth,trim={4.1cm 12.1cm 4.3cm 11.7cm},clip]{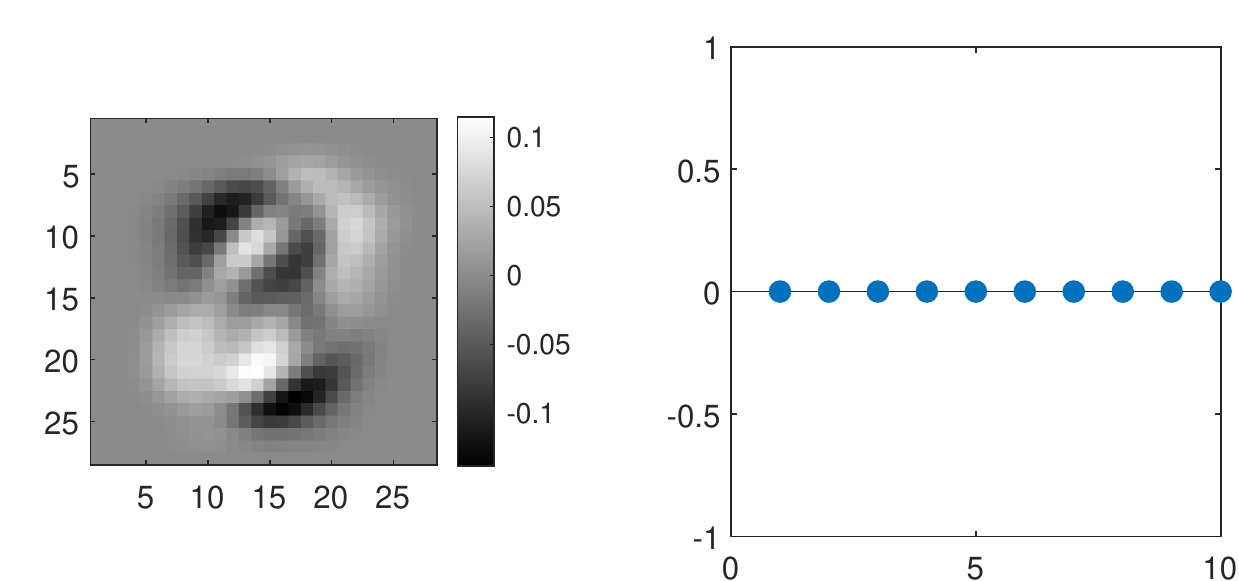}\\
\multicolumn{4}{|c|}{$\alpha=0$}\\
\hline
\hline
\includegraphics[width=.2\textwidth,trim={4.1cm 12.1cm 4.3cm 11.7cm},clip]{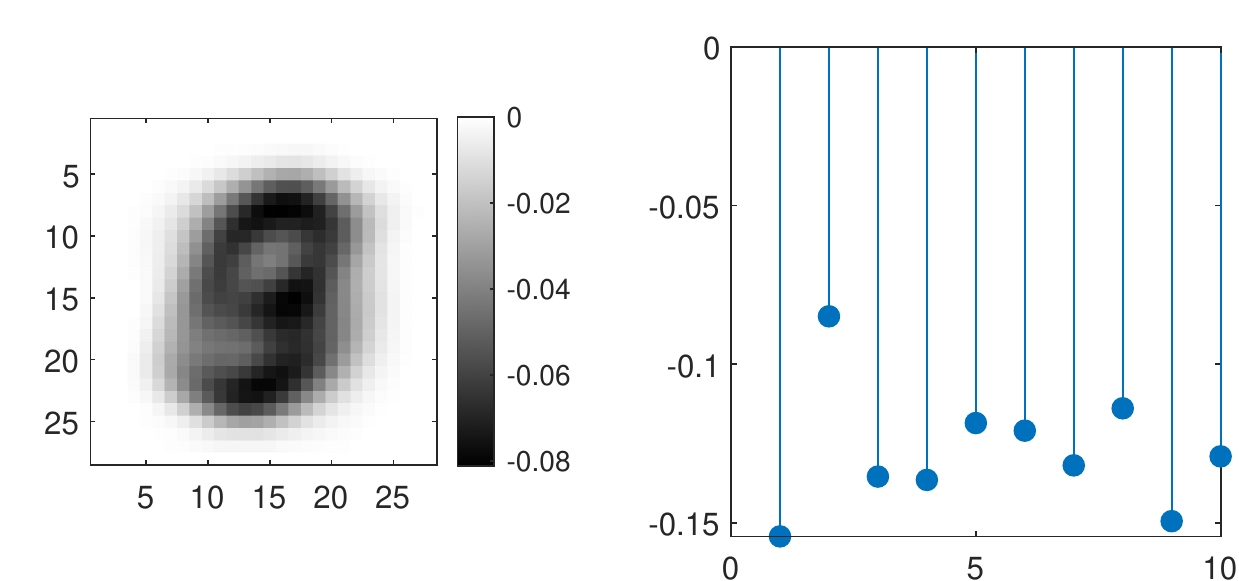}&
\includegraphics[width=.2\textwidth,trim={4.1cm 12.1cm 4.3cm 11.7cm},clip]{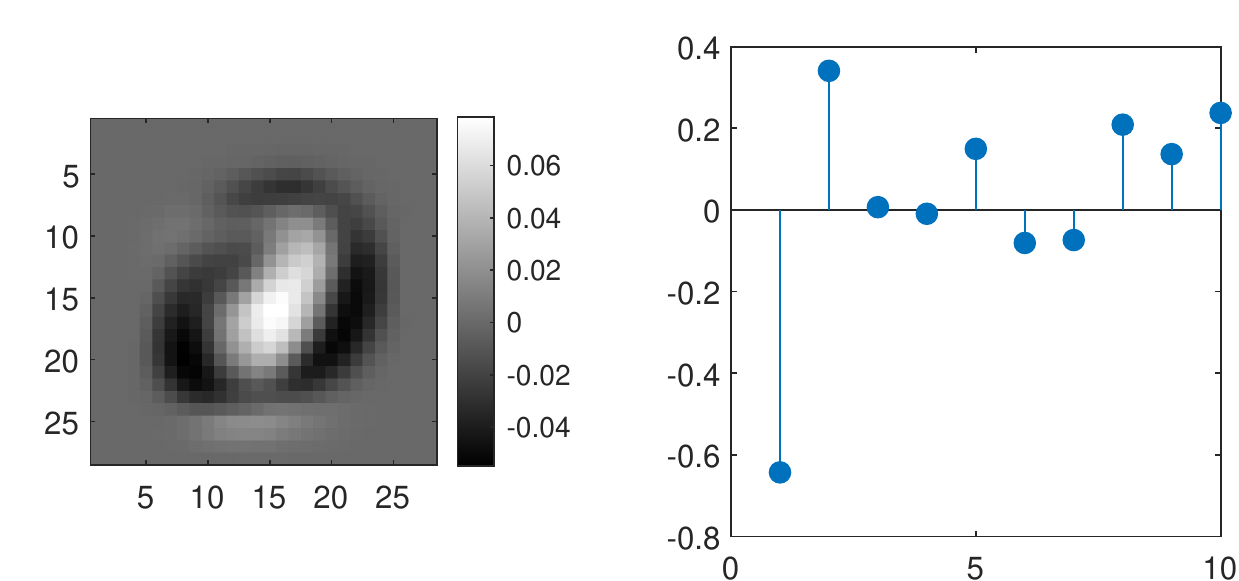}&
\includegraphics[width=.2\textwidth,trim={4.1cm 12.1cm 4.3cm 11.7cm},clip]{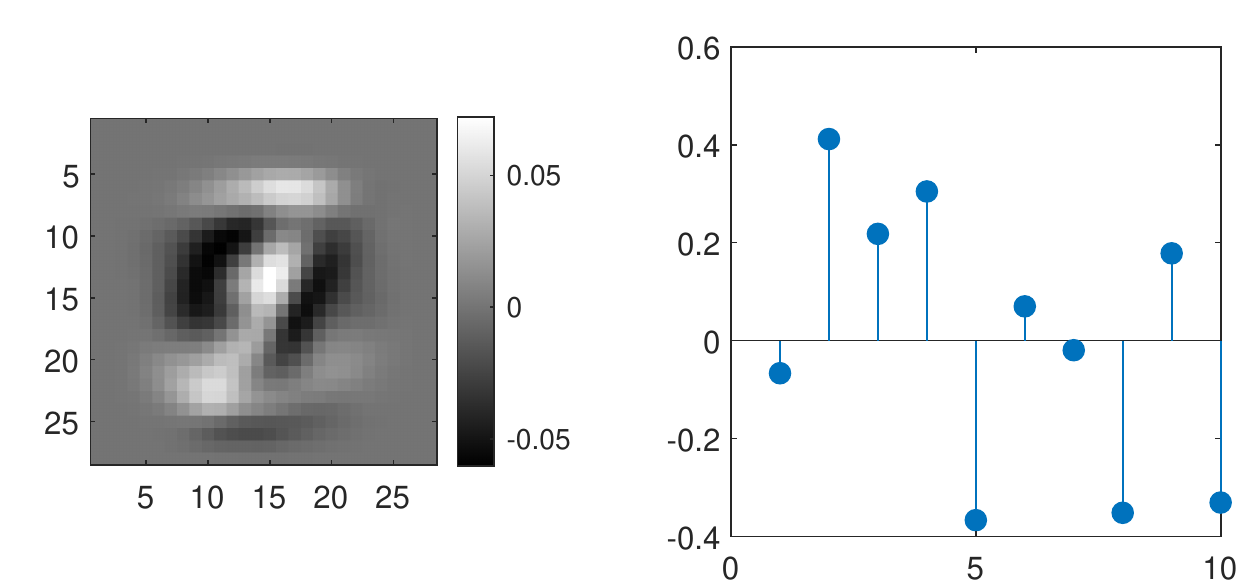}&
\includegraphics[width=.2\textwidth,trim={4.1cm 12.1cm 4.3cm 11.7cm},clip]{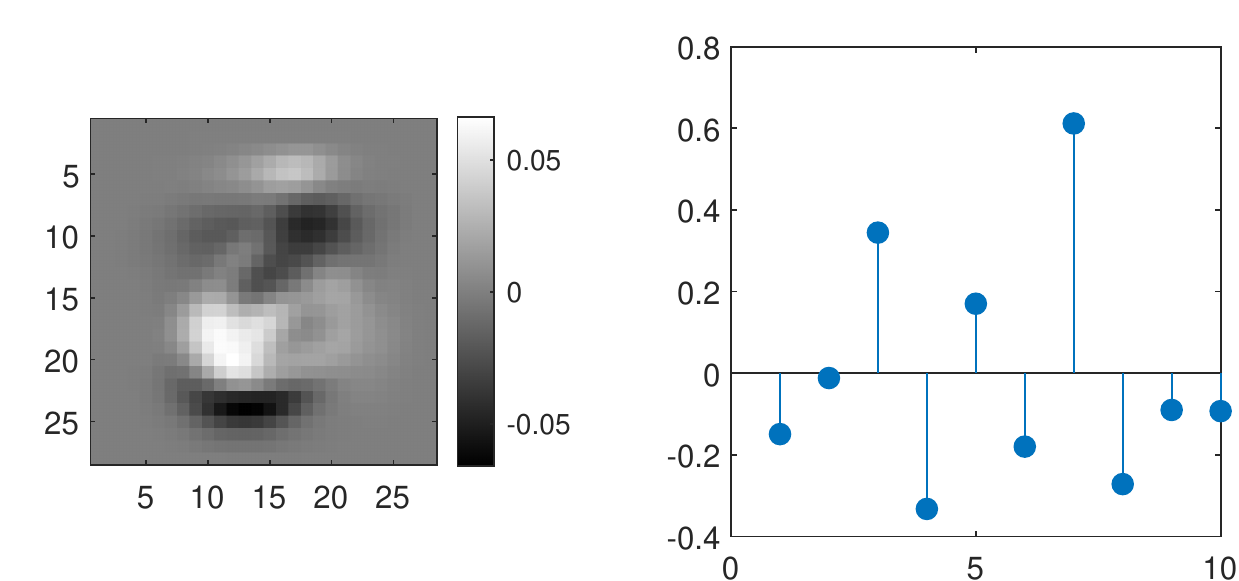}\\
\multicolumn{4}{|c|}{$\alpha=0.9$}\\
\hline
\end{tabular}
\caption{Principal components for classification computed  for Dataset MNIST. \footnotesize{Effect of $\alpha$ on the first 4 principal components, $\mathbf{u}_1$ to $\mathbf{u}_4$ (left to right), computed for $\alpha=0$ (top row) and $\alpha=0.9$ (bottom row). For visualisation, eigenvector $\mathbf{u}$ is shown as an image for the first $d_{\mathbf{x}}$ coordinates (in the feature $\mathbf{x}$-space) and as a plot for the remaining of its $n_c$ components (in the class $\mathbf{y}$-space). Data is not  centred (un-centered PCA) and the first component (left) captures the main deviation of the training dataset from the origin in the $\mathbf{z}$-space.}  }
\label{fig:principal:components}
\end{figure*}

\subsection{Data encoding with Class}

Class score vectors have recently been used as node attributes in a graph model for image segmentation \cite{ChopinICPRAI2022a}. We propose likewise to use that information explicitly by  creating a training dataset  noted $\mathcal{T}_{\alpha}=\lbrace \mathbf{z}_{\alpha}^{(i)}\rbrace_{i=1,\cdots,N}$  from the dataset $\mathcal{B}$, where each  instance $\mathbf{z}^{(i)}_{\alpha}$   concatenates the feature vector $\mathbf{x}^{(i)}$ with its class vector $\mathbf{y}^{(i)}$  as follow:
\begin{equation}
    \mathbf{z}_{\alpha}= (1-\alpha) \cdot
    \begin{pmatrix}
    \mathbf{x}\\
    \mathbf{0}_{\mathbf{y}}\\
    \end{pmatrix}
    +\alpha \cdot
     \begin{pmatrix}
    \mathbf{0}_{\mathbf{x}}\\
    \mathbf{y}\\
    \end{pmatrix}
    \label{eq:embedding:z}
\end{equation}
where    $\mathbf{0}_{\mathbf{x}}$ and $   \mathbf{0}_{\mathbf{y}}$ are the null vectors of feature space $\mathbb{R}^{d_{\mathbf{x}}}$  and class space $\mathbb{R}^{n_c}$  respectively. The scalar $0\leq\alpha \leq 1$ is  controlling the weight of the class vector w.r.t. the feature vector, and it is a hyper-parameter in this new framework. 
 The training dataset $\mathcal{T}_{\alpha}$ is stored in a data matrix noted $\mathrm{Z}_{\alpha}=\lbrack\mathbf{z}^{(1)}_{\alpha}, \cdots,  \mathbf{z}^{(N)}_{\alpha}\rbrack$.
 The matrix $\mathrm{Z}_{\alpha}$ concatenates vertically the matrix $\mathrm{X}$ and the matrix   
$\mathrm{Y}=\lbrack\mathbf{y}^{(1)}, \cdots,  \mathbf{y}^{(N)}\rbrack$ as follows:
\begin{equation}
\mathrm{Z}_{\alpha}=
\left\lbrack
\begin{array}{c}
(1-\alpha) \cdot    \mathrm{X}  \\
\alpha \cdot \mathrm{Y}\\
\end{array}\right\rbrack 
\end{equation}
We note $d_{\mathbf{z}}=d_{\mathbf{x}}+n_c$ the dimension of vectors $ \mathbf{z}_{\alpha}$, and the matrix  $\mathrm{Z}_{\alpha}$ is of size $d_{\mathbf{z}}\times N$.

\subsection{Principal components}

The $d_{\mathbf{z}}\times d_{\mathbf{z}}$ covariance matrix  $\Sigma_{\alpha}$ is computed as follow:  
\begin{equation}
  \Sigma_{\alpha}=\frac{1}{N}\ \mathrm{Z}_{\alpha}\mathrm{Z}_{\alpha}^T=\mathrm{U}_{\alpha}\Lambda_{\alpha}\mathrm{U}_{\alpha}^T
  \label{eq:cov:z}
\end{equation}
In our experiments, we  used Singular Value Decomposition (SVD) to compute  the diagonal matrix $\Lambda_{\alpha}$  of eigenvalues $\lbrace\lambda_i\rbrace_{i=1,\cdots,d_{\mathbf{z}}}$ of $\Sigma_{\alpha}$ and with the corresponding eigenvectors stored as columns in the matrix $\mathrm{U}_{\alpha}=[\mathbf{u}_{ 1},\cdots,\mathbf{u}_{d_{\mathbf{z}}}]$. 
For large training dataset ($N>>0)$,  more efficient algorithms alternative  to SVD  can be used to compute the principal components \cite{gemp2021eigengame}.
Eigenvalues are sorted in decreasing order $\lambda_1\geq \lambda_2\cdots\geq \lambda_{d_{\mathbf{z}}}\geq 0$ and their associated eigenvectors form a orthonormal basis of the $\mathbf{z}$-space.
When $\alpha=0$, only feature vectors appear in the vector $\mathbf{z}_{0}$ corresponding to the standard usage of PCA for dimensionality reduction of the  feature  space. When $\alpha>0$, the principal components stored in $\mathrm{U}_{\alpha}$
change away from the baseline $\mathrm{U}_{0}$ (e.g. see Fig. \ref{fig:principal:components}).

\subsection{Encoder \& Decoder Model $\mathcal{M}^{n_e}_{\alpha}$}

\begin{figure}[!h]
\begin{center}
\begin{tabular}{cc}
\includegraphics[width=.45\linewidth,trim={4.1cm 10cm 4.2cm 9.7cm},clip]{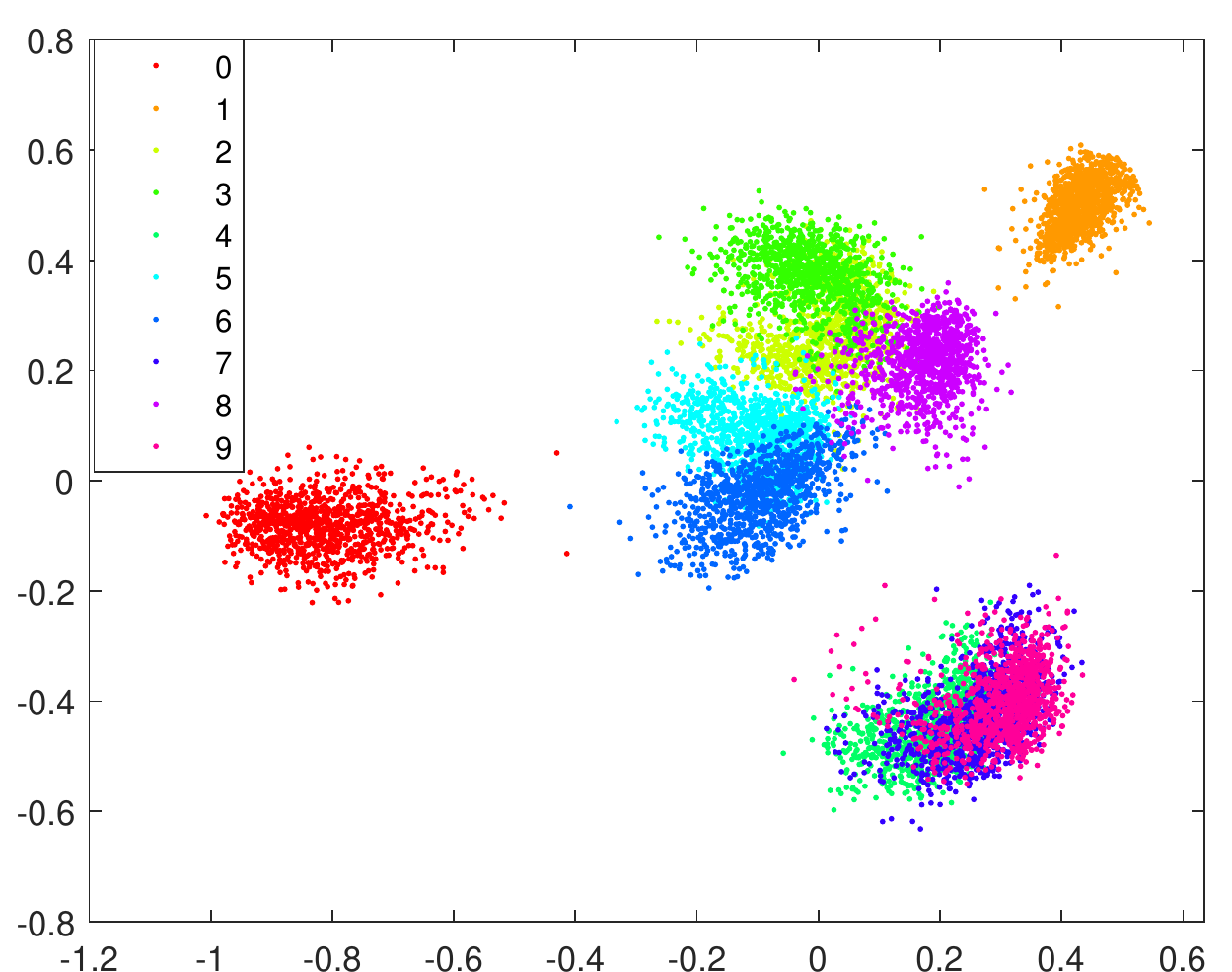}&
\includegraphics[width=.45\linewidth,trim={4.1cm 10cm 4.2cm 9.7cm},clip]{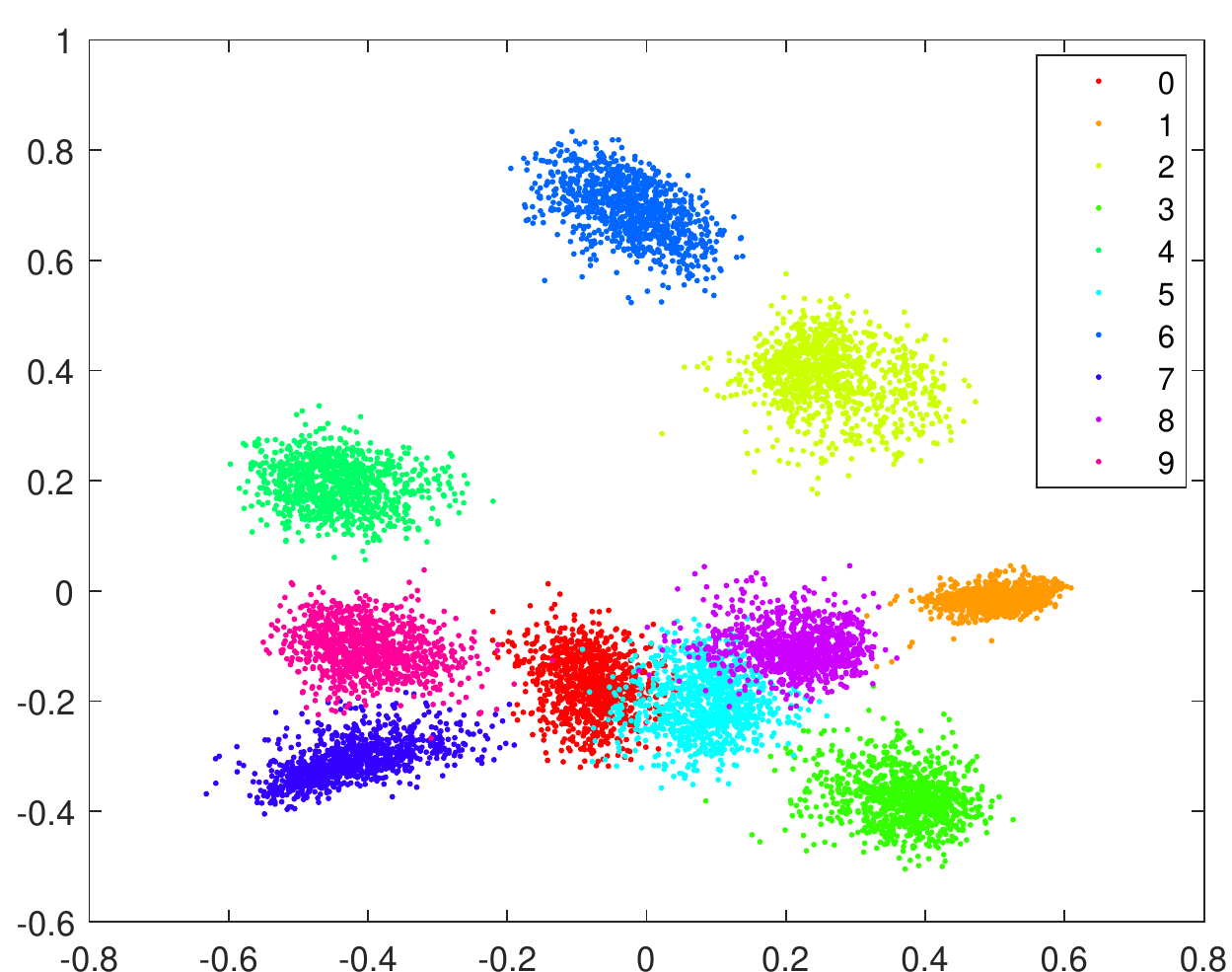}\\
\footnotesize{$(\mathbf{u}_2,\mathbf{u}_3)$} &  \footnotesize{$(\mathbf{u}_3,\mathbf{u}_4)$}\\
\end{tabular}
\end{center}
\caption{Visualisation of projections. \footnotesize{The training set $\mathcal{T}_{\alpha=0.9}$ is projected in the eigenspaces defined by pairs of our learnt principal components for dataset MNIST (shown Fig. \ref{fig:principal:components}).}}
\label{fig:MNIST10:TSNE}
\end{figure}

Dimensionality reduction is  performed by only considering the first $n_e$ eigenvectors associated with the $n_e$ highest eigenvalues. Noting $\mathrm{U}_{\alpha}^{n_e}=[\mathbf{u}_{ 1},\cdots,\mathbf{u}_{n_e}]$, the projection of any vector $\mathbf{z}$ against the first $n_e$ eigenvectors is computed as:
\begin{equation}
\mathbf{p}_{\alpha}^{n_e}=\left(\mathrm{U}_{\alpha}^{n_e}\right)^T \mathbf{z}   
\label{eq:encoder}
\end{equation}
 Projections of the training set  $\mathcal{T}_{\alpha}$ can be visualised  in the eigenspaces defined by a pair of principal components (cf. Fig. \ref{fig:MNIST10:TSNE}) providing renderings similar to  TSNE \cite{JMLR:v9:vandermaaten08a}.

\begin{figure}[!h]
\includegraphics[width=\linewidth]{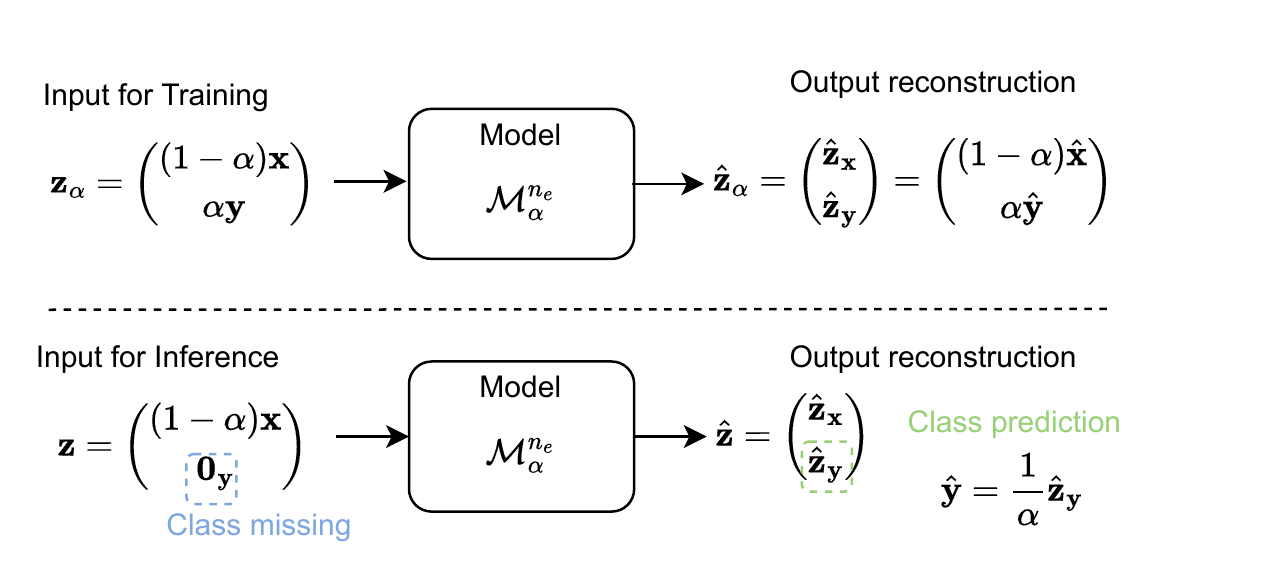}
\caption{Using model $\mathcal{M}^{n_e}_{\alpha}$: \footnotesize{At training stage, hyper-parameters $\alpha$ and $n_e$ are chosen, principal components $\mathrm{U}_{\alpha}^{n_e}$ are computed as parameters for model $\mathcal{M}^{n_e}_{\alpha}$. Note that the reconstruction error $\mathbb{E}_{\mathcal{T}_{\alpha}}[\|\mathbf{z}_{\alpha}-\hat{\mathbf{z}_{\alpha}} \|]$ is minimized when using $\mathrm{U}_{\alpha}^{n_e}$. At test time, the input used is proportional to $\mathbf{z}_0$ and a class prediction appears as part of the output reconstruction. }}
\label{fig:machine:usage}
\end{figure}

\begin{figure*}[t]
    \centering
    \begin{tabular}{|ccc|}
    \hline
    \includegraphics[width=.3\linewidth,trim={4.1cm 10cm 4.2cm 9.7cm},clip]{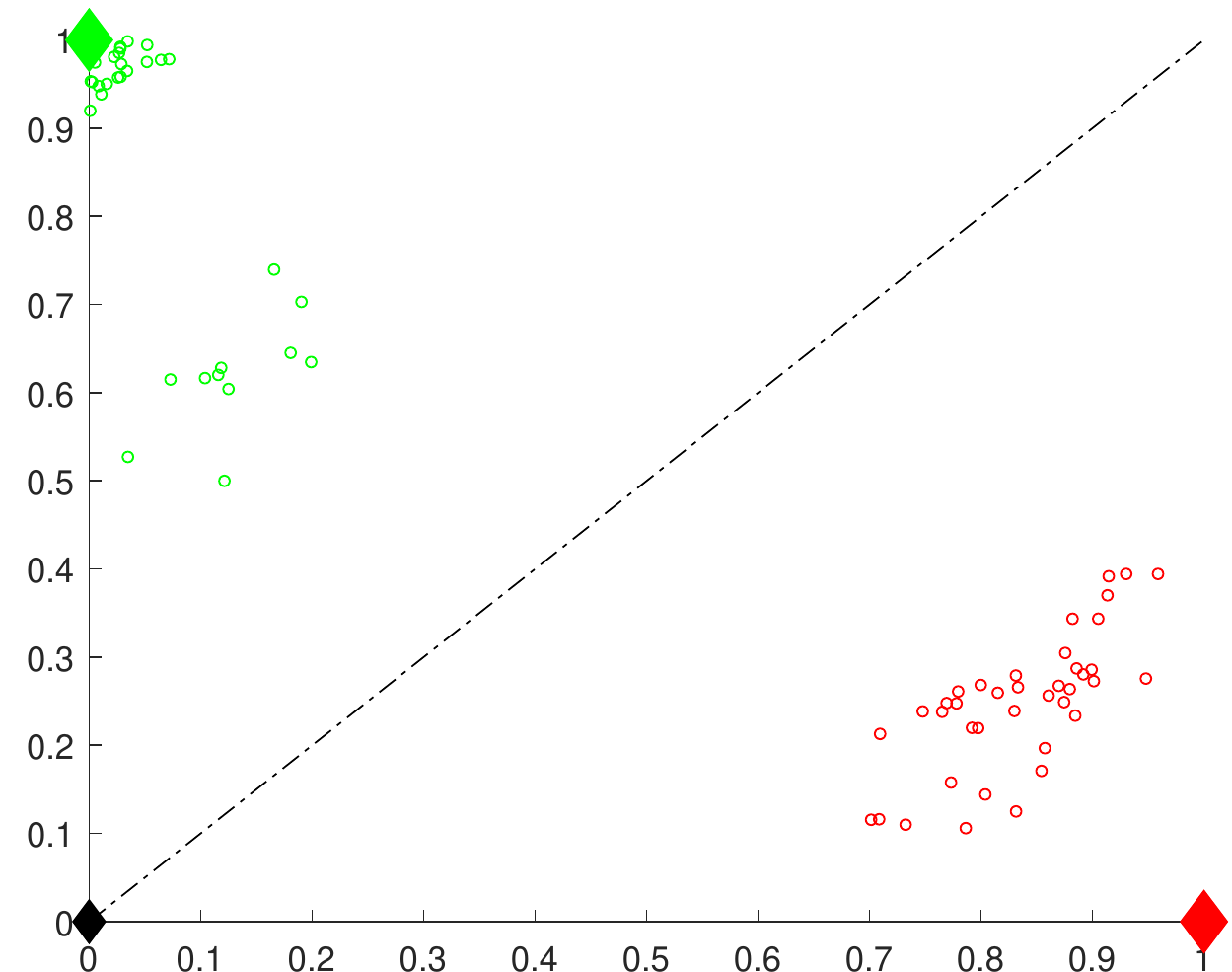}&
    \includegraphics[width=.3\linewidth,trim={4.1cm 10cm 4.2cm 9.7cm},clip]{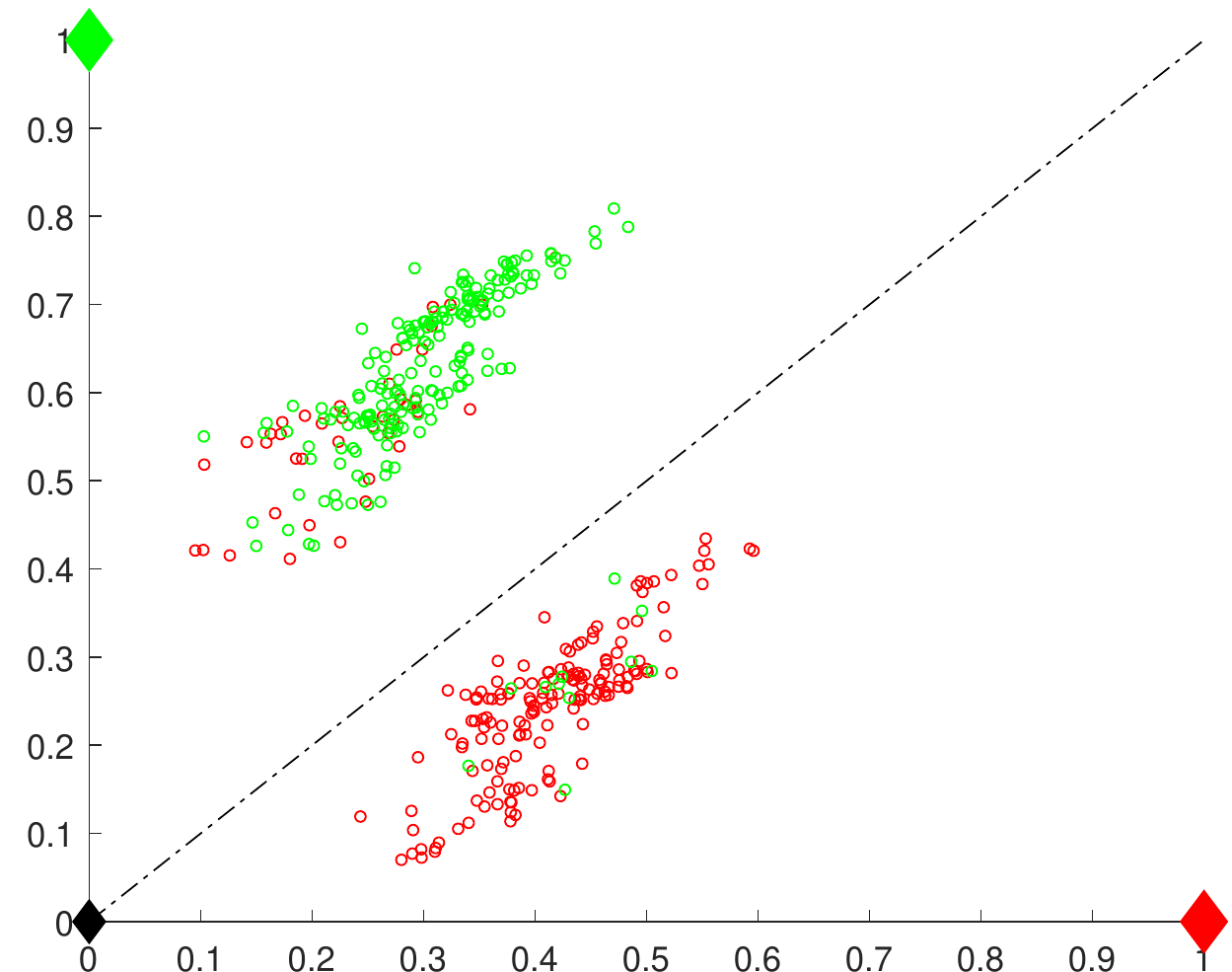}&
    \includegraphics[width=.3\linewidth,trim={4.1cm 10cm 4.2cm 9.7cm},clip]{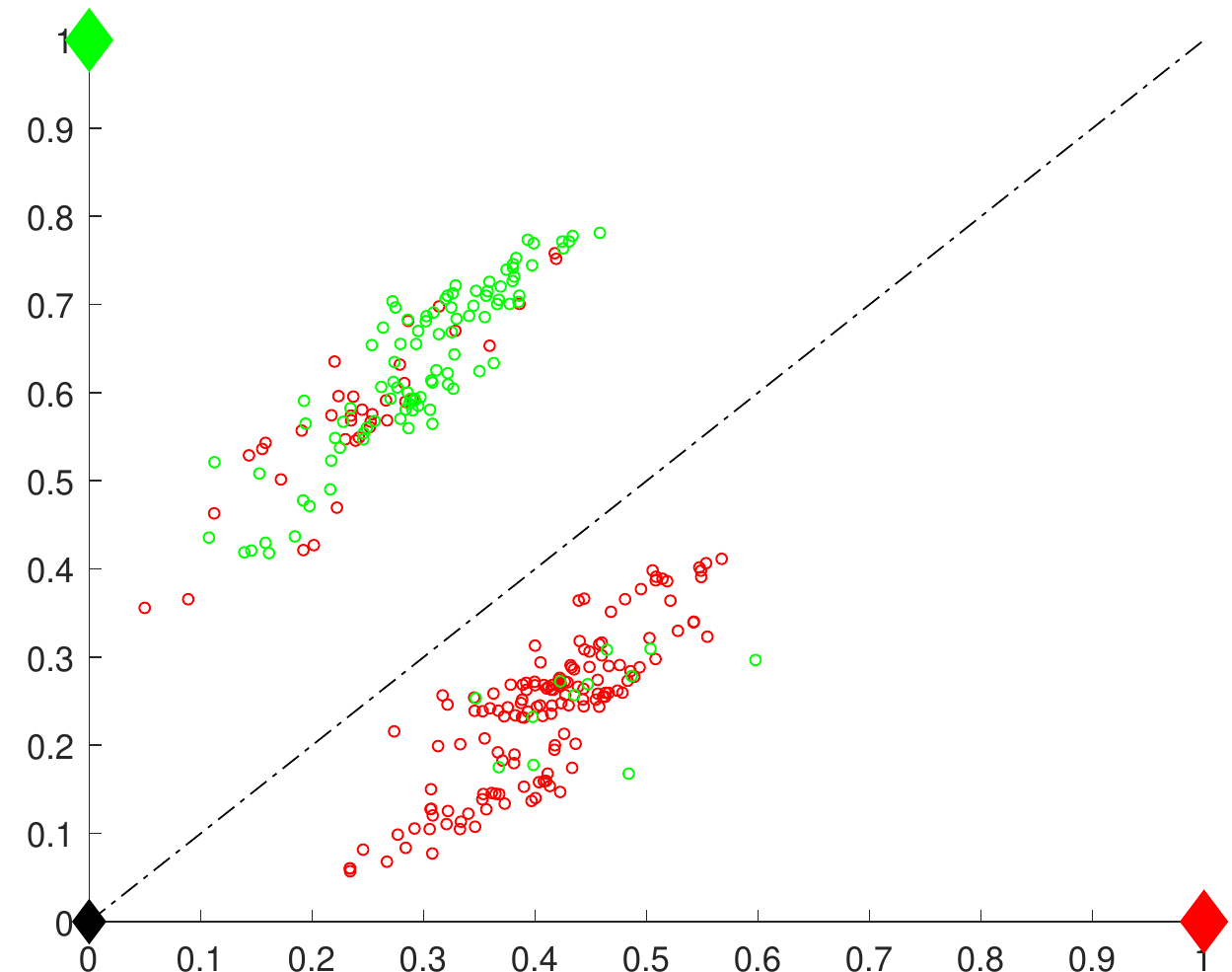}
    \\
    $\mathcal{I}^{\mathbf{x}}_{\mathbf{y}}$: acc=1 &
 $\mathcal{I}^{\mathbf{x}}_{0}$: acc=0.8725 &  
 $\mathcal{I}^{\mathbf{x'}}_{0}$:   acc=0.8310 \\
  \hline
 \end{tabular}
    \caption{Visualisation in the $\mathbf{y}$-space. \footnotesize{Dataset \textit{australian}: visualisation  of the outputs  $\hat{\mathbf{y}}$  (dots colour coded with the ground truth class label)  in the class  $\mathbf{y}$-space of dimension $n_c=2$ computed with model $\mathcal{M}^{n_e=4}_{\alpha=0.5}$.   The black diamond shows the class input to our model for sets  $\mathcal{I}^{\mathbf{x}}_{0}$ and $\mathcal{I}^{\mathbf{x'}}_{0}$; the red and green diamonds are the ideal class vectors used as part of the input in set $\mathcal{I}^{\mathbf{x}}_{\mathbf{y}}$ (cf. Sec. \ref{sec:input:sets}).}}
    \label{fig:move}
\end{figure*}

Choosing a small number $n_e$ of eigenvectors for projection  allows to limit the number of computations, but it also needs to be large enough to capture useful information about the dataset $\mathcal{T}_{\alpha}$. The projection (Eq.\ref{eq:encoder})
is a linear encoder   of $\mathbf{z}$ which can be decoded as follows:
\begin{equation}
\hat{\mathbf{z}} = \mathrm{U}_{\alpha}^{n_e} \mathbf{p}_{\alpha}^{n_e}=\mathrm{U}_{\alpha}^{n_e}\left(\mathrm{U}_{\alpha}^{n_e}\right)^T \mathbf{z}   
\label{eq:decoder}
\end{equation}
Our model $\mathcal{M}^{n_e}_{\alpha}$  is now explicitly defined (Eq. (\ref{eq:decoder})) for transforming any input $\mathbf{z}$ for transformation into an output $\hat{\mathbf{z}}$, using  the learnt parameters  (matrix $\mathrm{U}^{n_e}_{\alpha}$) for the chosen hyper-parameters $\alpha$ and $n_e$  (cf. Fig. \ref{fig:machine:usage}).  
From the output $\hat{\mathbf{z}}$, the subvector $\hat{\mathbf{y}}$ is analysed such that the class $j\in\lbrace 1,\cdots,n_c\rbrace$ is identified by locating the highest valued coordinate of vector $\hat{\mathbf{y}}$:
\begin{equation}
\hat{j}=\arg\max_{j=1,\cdots,n_c}\left\lbrace \hat{\mathbf{y}}^T=[\hat{y}_1,\cdots,\hat{y}_j,\cdots,\hat{y}_{n_c}]\right\rbrace
\label{eq:inference}
\end{equation}
 Estimates $\hat{\mathbf{y}}$ can also be visualised in the $\mathbf{y}$-space (cf. Fig. \ref{fig:move}).

\section{Experimental results}
\label{sec:input:sets}

We rescale feature vectors  $ \mathbf{x}$ so that the $\mathbf{x}$-space is an hypercube of volume 1.
For  MNIST dataset,  all pixel values are scaled between 0 and 1 by dividing by 255. 
For the \textit{wine} and \textit{australian} datasets, all feature dimensions are  re-scaled to 1 by dividing by their corresponding maximum found in the dataset. 
For the \textit{australian} dataset, we have used  200 instances per class for creating a balanced training set $\mathcal{B}$ ($N=2\times200=400$)
and the remaining $N'=290$  instances are used  for  the set $\mathcal{B}'$ (a random allocation of instances is performed between $\mathcal{B}$ and $\mathcal{B}'$).
Likewise for the  \textit{wine} dataset, 40 instances per class are used for creating a balanced training set $\mathcal{B}$ ($N=3\times40=120$) and the rest $N'=58$ is used for the test set $\mathcal{B}'$ (random allocation is performed at every run).
For the MNIST dataset, 1000 instances per class is used for training ($N=10\times 1000=10000$) and a balanced test set is likewise defined with  $N'=10\times 1000=10000$ exemplars. 

   From the training set $\mathcal{B}=\lbrace(\mathbf{x}^{(i)},\mathbf{y}^{(i)})\rbrace_{i=1,\cdots,N}$, the set $\mathcal{T}_{\alpha}=\lbrace \mathbf{z}_{\alpha}^{(i)} \rbrace$ is used to compute the  matrix of eigenvectors $\mathrm{U}_{\alpha}$. It is also used as a set of inputs noted $\mathcal{I}^{\mathbf{x}}_{\mathbf{y}}\equiv\mathcal{T}_{\alpha}$ for testing our model $\mathcal{M}^{n_e}_{\alpha}$:  both feature vectors and class vectors are available as seen during training of the model $\mathcal{M}^{n_e}_{\alpha}$ therefore  we expect accuracy reported for $\mathcal{I}^{\mathbf{x}}_{\mathbf{y}}$  to be high if not perfect when  the number $n_e$ of eigenvectors is large enough to prevent any loss of information.  In practice the class vector is not known and only the feature vector is available to be used as input of the our model $\mathcal{M}^{n_e}_{\alpha}$ (cf. Fig. \ref{fig:machine:usage}). We propose to use the input $\mathbf{z}_0$ (multiply by $(1-\alpha)$, cf. Eq. \ref{eq:embedding:z}) that initialises all components in the class vector to 0s. 
    From the training set $\mathcal{B}=\lbrace(\mathbf{x}^{(i)},\mathbf{y}^{(i)})\rbrace_{i=1,\cdots,N}$, we create a test set of inputs noted $\mathcal{I}_{0}^{\mathbf{x}}=\lbrace  (1-\alpha)\mathbf{z}_{0}^{(i)} \rbrace$ that have the same feature vectors as seen during training by the model but without its class information.
    From  $\mathcal{B}'$ 
    unseen for computing the  matrix $\mathrm{U}_{\alpha}$, we create likewise a set of inputs noted  $\mathcal{I}_{\mathbf{0}}^{\mathbf{x'}}$  to test our model for classification.
For each set of inputs $\mathcal{I}^{\mathbf{x}}_{\mathbf{y}}$, $\mathcal{I}^{\mathbf{x}}_{\mathbf{0}}$ and $\mathcal{I}^{\mathbf{x'}}_{\mathbf{0}}$, 
our model is used to estimate the class label (cf. Eq. \ref{eq:inference})  and we report the accuracy rate.   

\begin{figure*}[!t]
    \centering
    \begin{tabular}{|ccc|}
    \hline
      $\mathcal{I}^{\mathbf{x}}_{\mathbf{y}}$ & 
 $\mathcal{I}^{\mathbf{x}}_{0}$   & 
 $\mathcal{I}^{\mathbf{x'}}_{0}$   \\
        \hline
         \hline
    \multicolumn{3}{|p{.95\linewidth}|}{\footnotesize{Dataset \textit{australian}: $n_c=2$, $N=n_c\times 200$, $N'=290$, $d_{\mathbf{x}}=14$, $d_{\mathbf{z}}=16$. Accuracy results reported in \cite{10.5555/3045390.3045650}  using the entire 14 dimensional feature space, range  in between 0.65  to 0.85. Our model   $\mathcal{M}_{0.2}^{4}$ has  average classification accuracy rates ($\pm$standard deviations) of $(0.81\pm0.03;0.77 \pm0.02;0.76\pm0.02)$ for sets $(\mathcal{I}^{\mathbf{x}}_{\mathbf{y}};\mathcal{I}^{\mathbf{x}}_{0};\mathcal{I}^{\mathbf{x'}}_{0})$  computed over 10 runs. With an additional principal component, our models $\mathcal{M}_{0.2}^{5}$ and $\mathcal{M}_{0.4}^{5}$ have classification accuracy rates  $(0.86 \pm0.01;	0.86  \pm0.01; 0.84  \pm0.02)$  and $(0.98 \pm0.007; 0.86 \pm0.01 ;	0.84\pm0.02)$ respectively.  }}\\
\includegraphics[width=.3\linewidth,trim={4.1cm 10cm 4.4cm 9.7cm},clip]{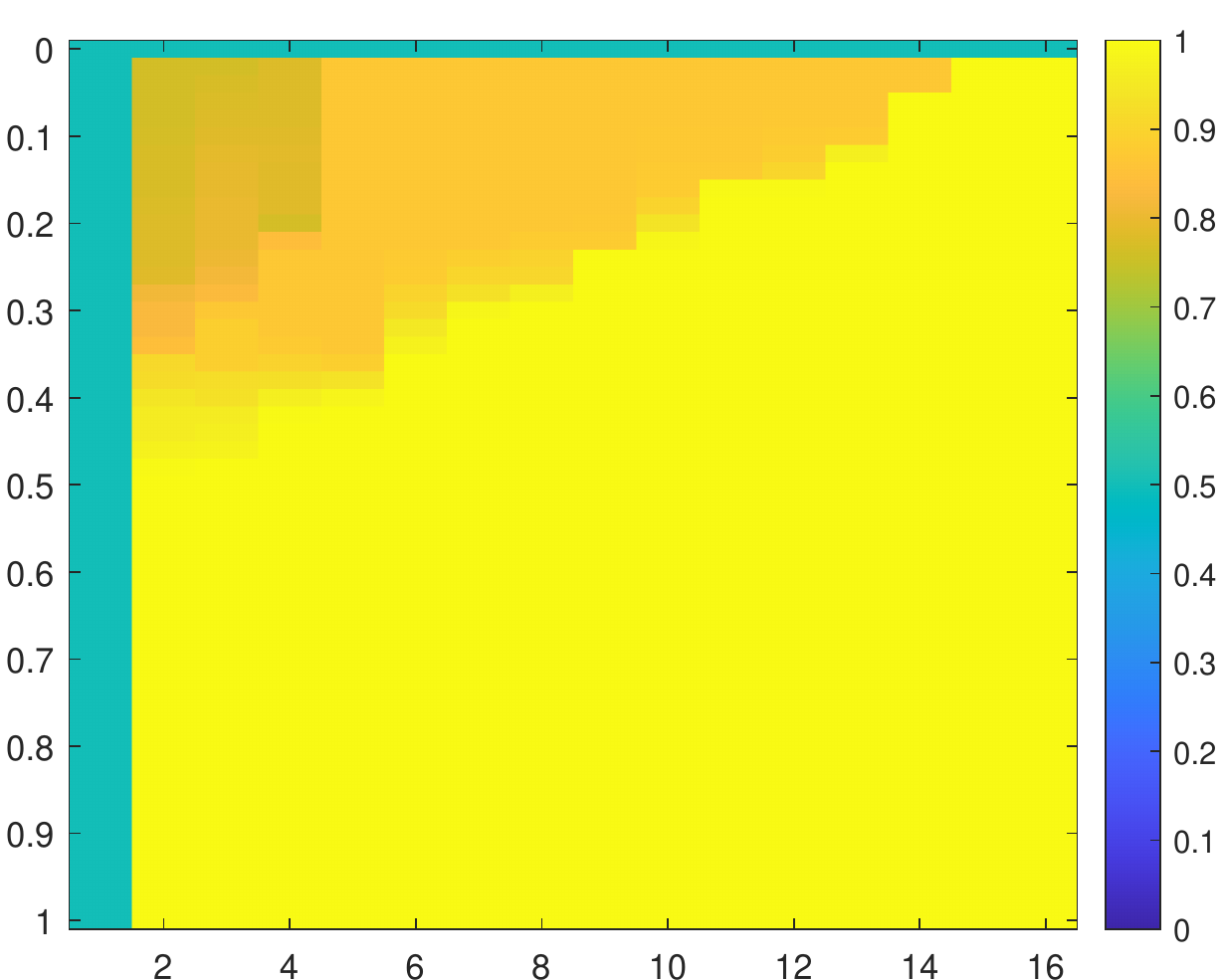}  &
\includegraphics[width=.3\linewidth,trim={4.1cm 10cm 4.4cm 9.7cm},clip]{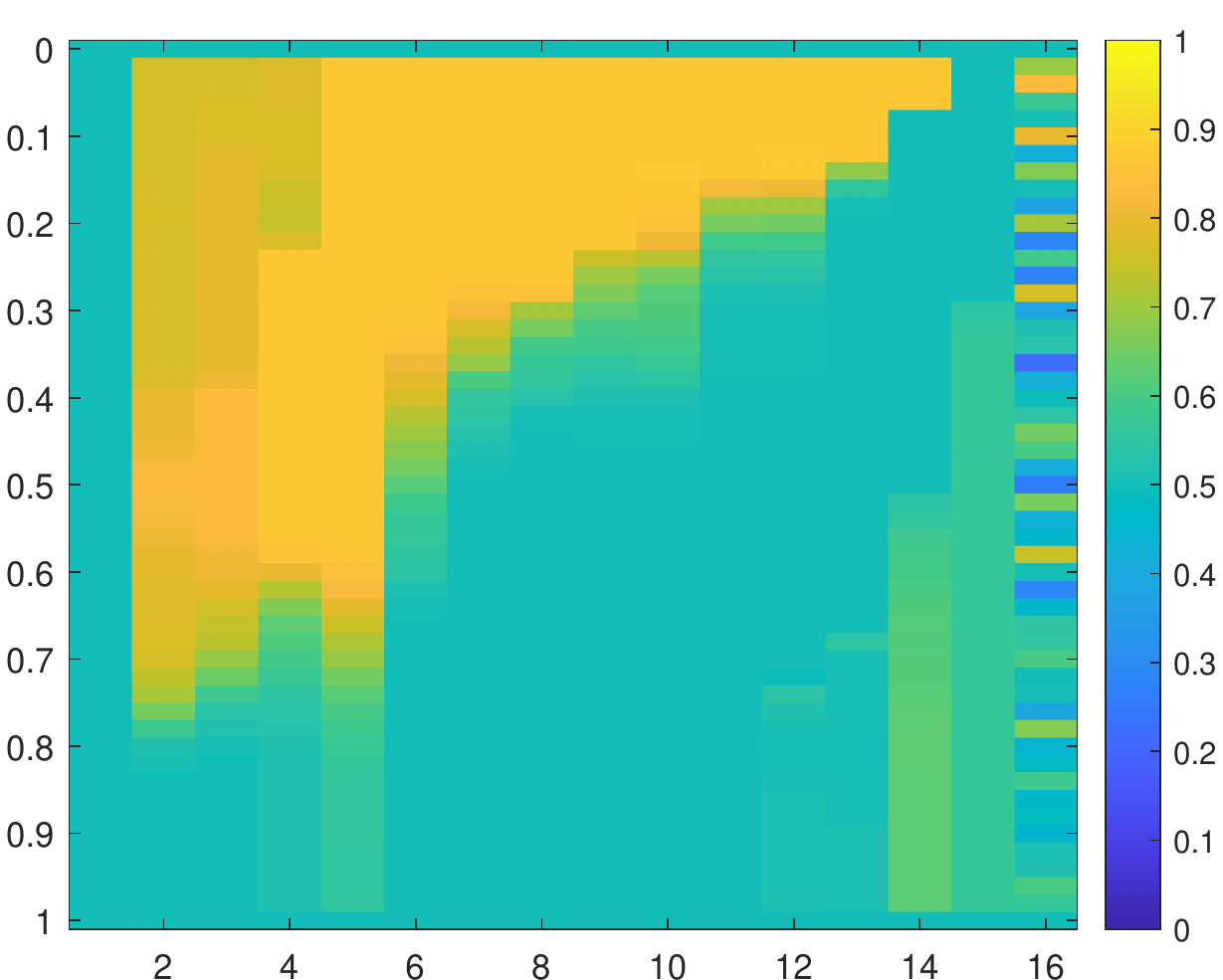} &
\includegraphics[width=.3\linewidth,trim={4.1cm 10cm 4.4cm 9.7cm},clip]{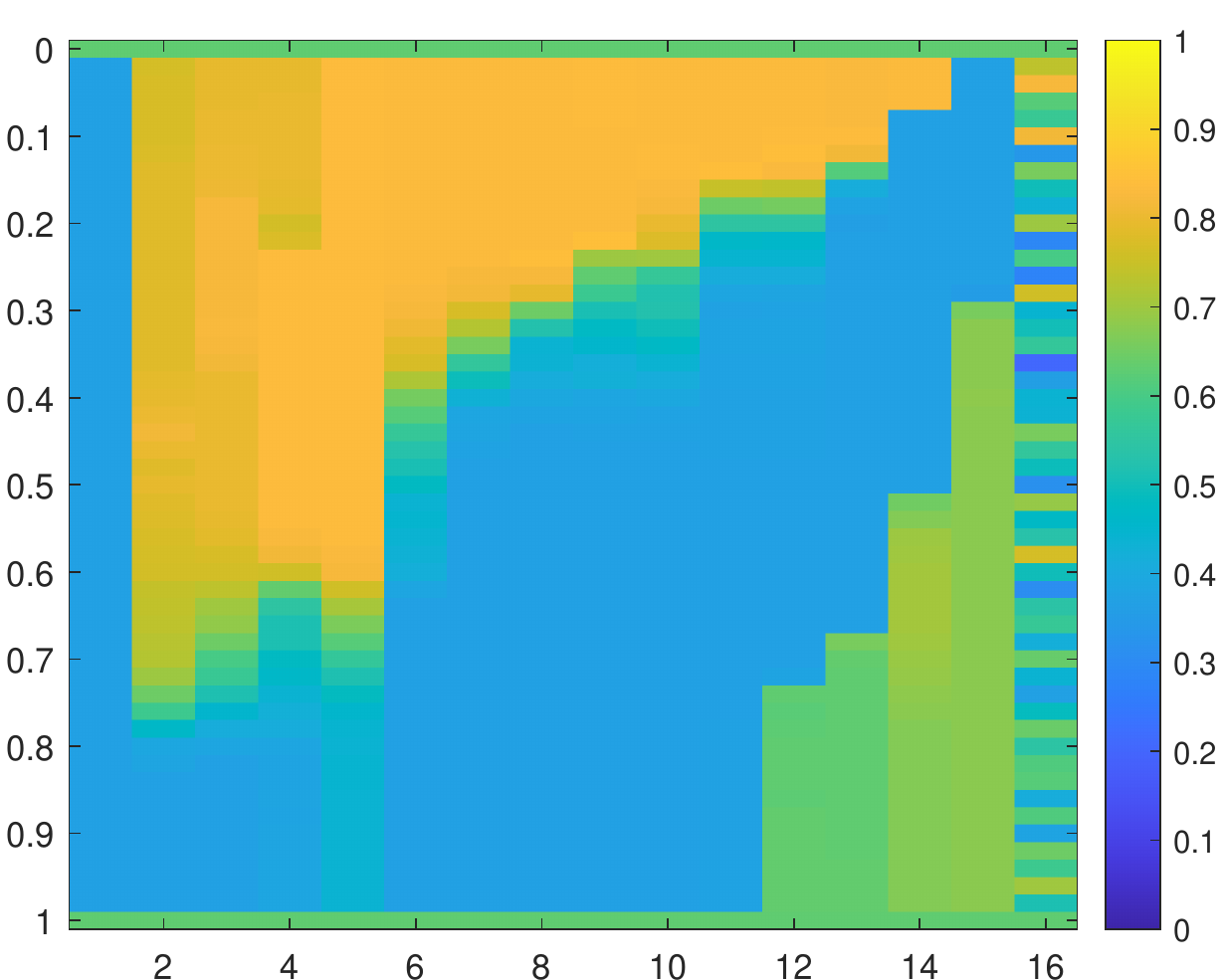} \\

\hline
    \hline
    \multicolumn{3}{|p{.95\linewidth}|}{\footnotesize{Dataset \textit{wine}: $n_c=3$, $N=n_c\times 40$, $N'=58$, $d_{\mathbf{x}}=13$, $d_{\mathbf{z}}=16$. Accuracy results  reported in \cite{JMLR:v9:vandermaaten08a,10.5555/3045390.3045650,Eusipco2022Collas}  using the entire 13 dimensional feature space, range in between 0.70  to 0.98. Our model   $\mathcal{M}_{0.2}^{4}$ has average classification accuracy rates ($\pm$standard deviations) of $(0.99\pm0.005;0.94 \pm0.02;0.88\pm0.03)$ for sets $(\mathcal{I}^{\mathbf{x}}_{\mathbf{y}};\mathcal{I}^{\mathbf{x}}_{0};\mathcal{I}^{\mathbf{x'}}_{0})$ computed over 10 runs. With an additional principal component, our models $\mathcal{M}_{0.2}^{5}$ and $\mathcal{M}_{0.4}^{5}$ have classification accuracy rates  $(0.99\pm0.003);0.96\pm0.01;0.92\pm0.02)$  and $(1.00 \pm 0; 0.95\pm0.02;0.91\pm0.02)$ respectively.}}\\
\includegraphics[width=.3\linewidth,trim={4.1cm 10cm 4.4cm 9.7cm},clip]{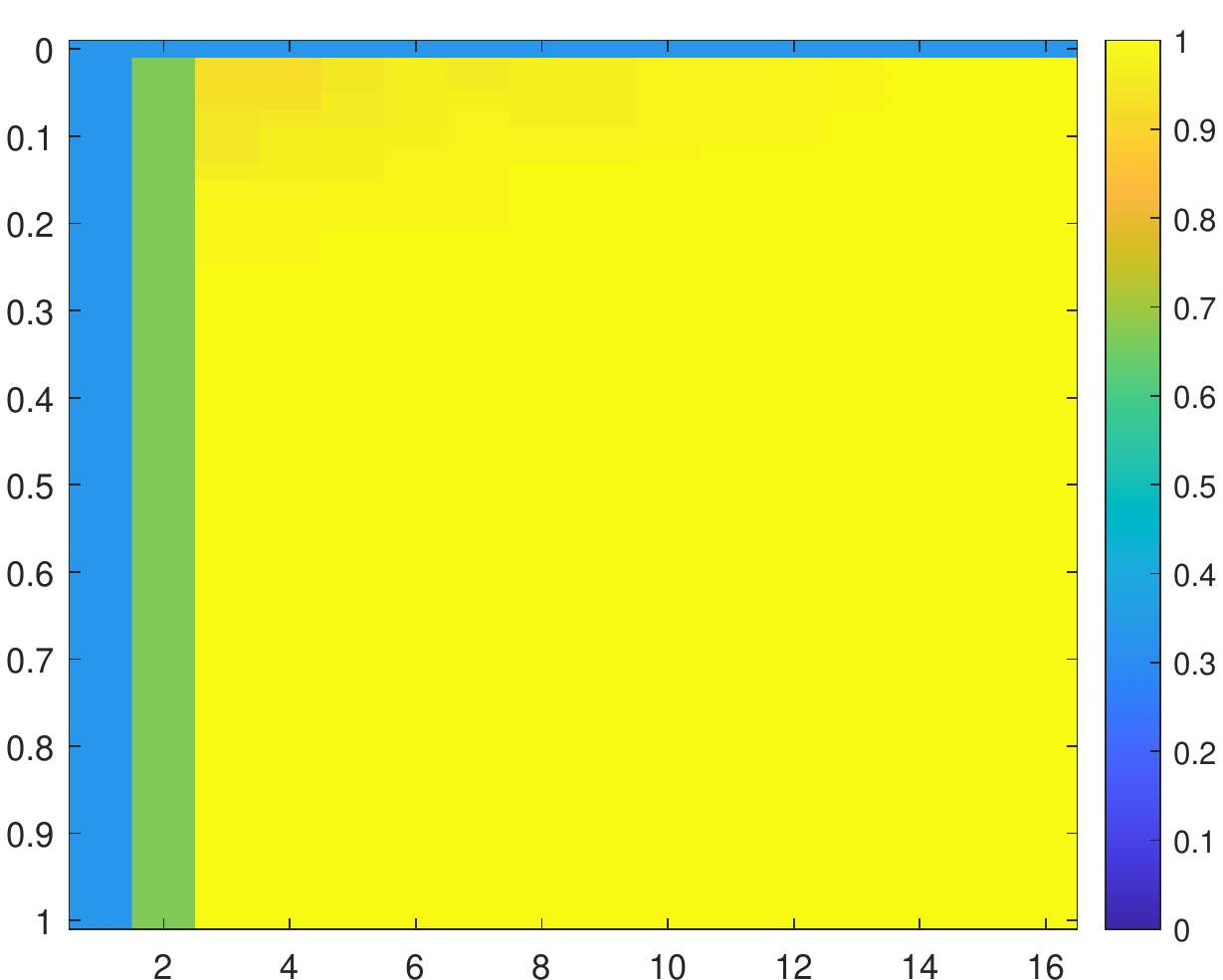}  &
\includegraphics[width=.3\linewidth,trim={4.1cm 10cm 4.4cm 9.7cm},clip]{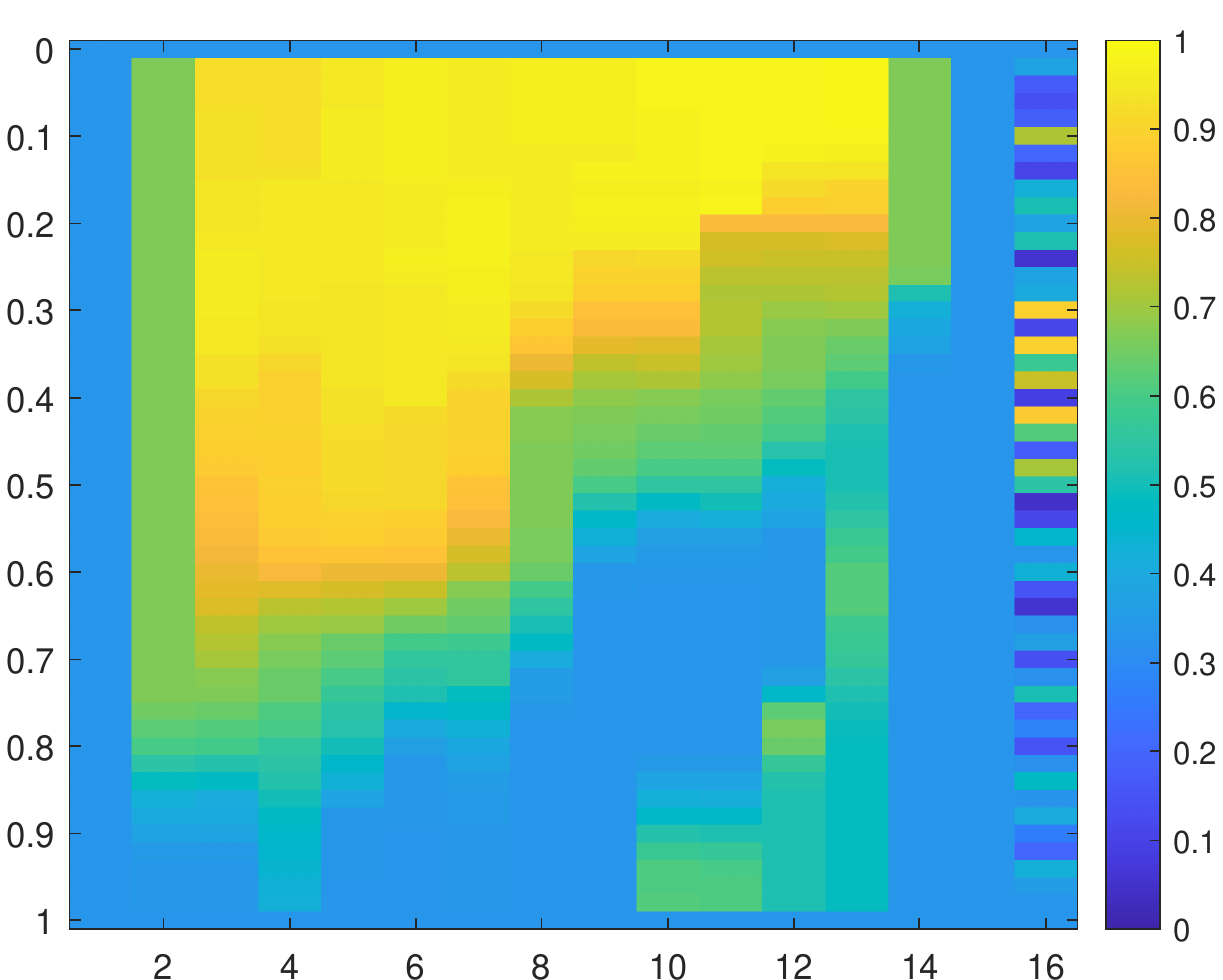} &
\includegraphics[width=.3\linewidth,trim={4.1cm 10cm 4.4cm 9.7cm},clip]{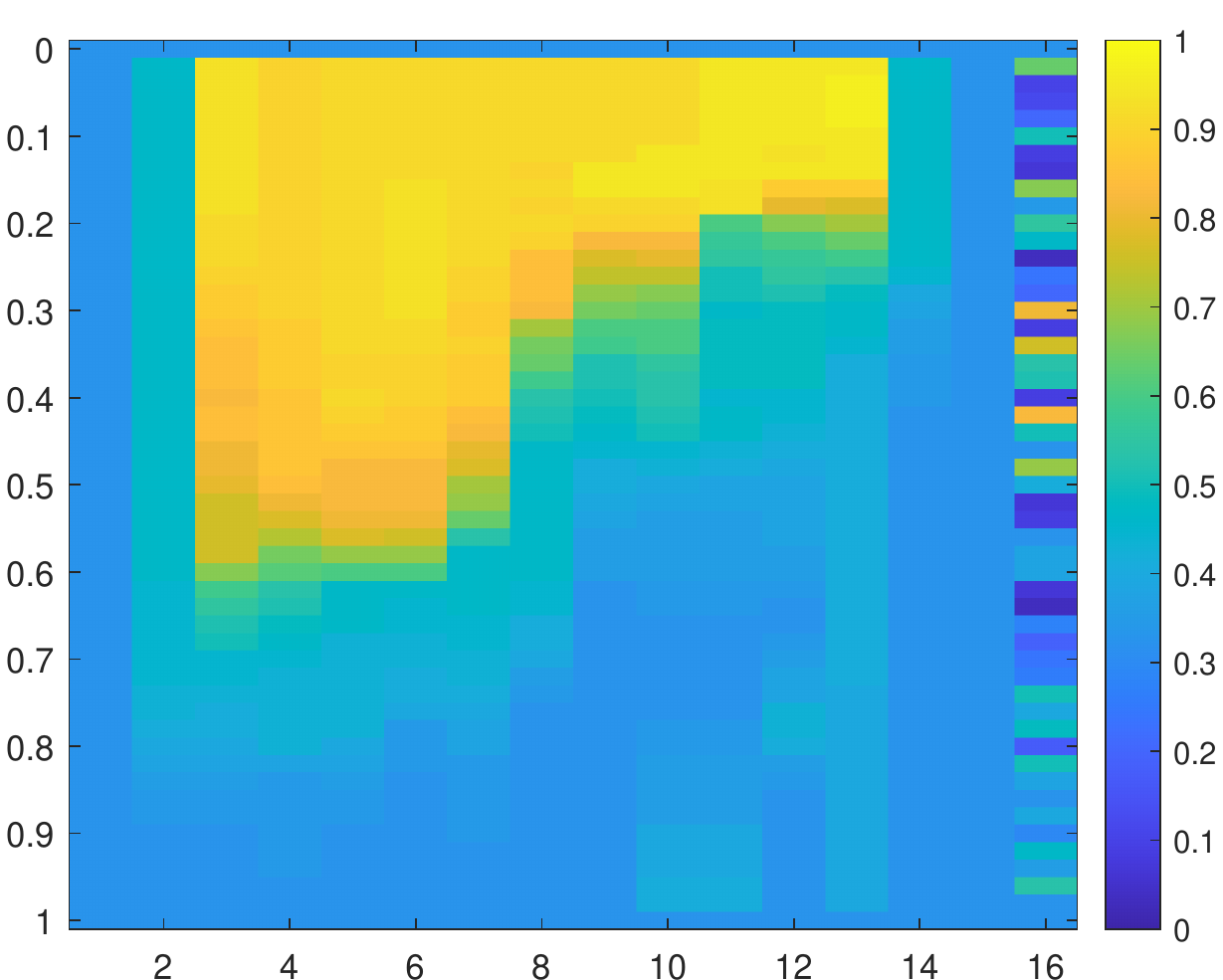} \\
\hline
\hline
    \multicolumn{3}{|p{.95\linewidth}|}{\footnotesize{Dataset \textit{MNIST}: $n_c=10$, $N=n_c\times 1000$, $N'=n_c\times 1000$, $d_{\mathbf{x}}=784$, $d_{\mathbf{z}}=794$.   \cite{10.5555/1577069.1577078} reports accuracy ranging between 0.93 to 0.99 in reduced space of 164 dimensions, \cite{10.5555/3045390.3045650} reports accuracy ranging between 0.82 to 0.88   using the entire 784 dimensional feature space dimensions. 
    Our model $\mathcal{M}_{0.2}^{164}$ has accuracy rates $(0.8661 ; 0.8598 ; 	0.8247)$  for sets $(\mathcal{I}^{\mathbf{x}}_{\mathbf{y}};\mathcal{I}^{\mathbf{x}}_{0};\mathcal{I}^{\mathbf{x'}}_{0})$. With only 16 principal components,  our model $\mathcal{M}_{0.9}^{16}$ has also a good performance $(1.00; 0.8089;  0.7883)$ for a much lower computational cost.   }}\\
\includegraphics[width=.3\linewidth,trim={4.1cm 10cm 4.4cm 9.7cm},clip]{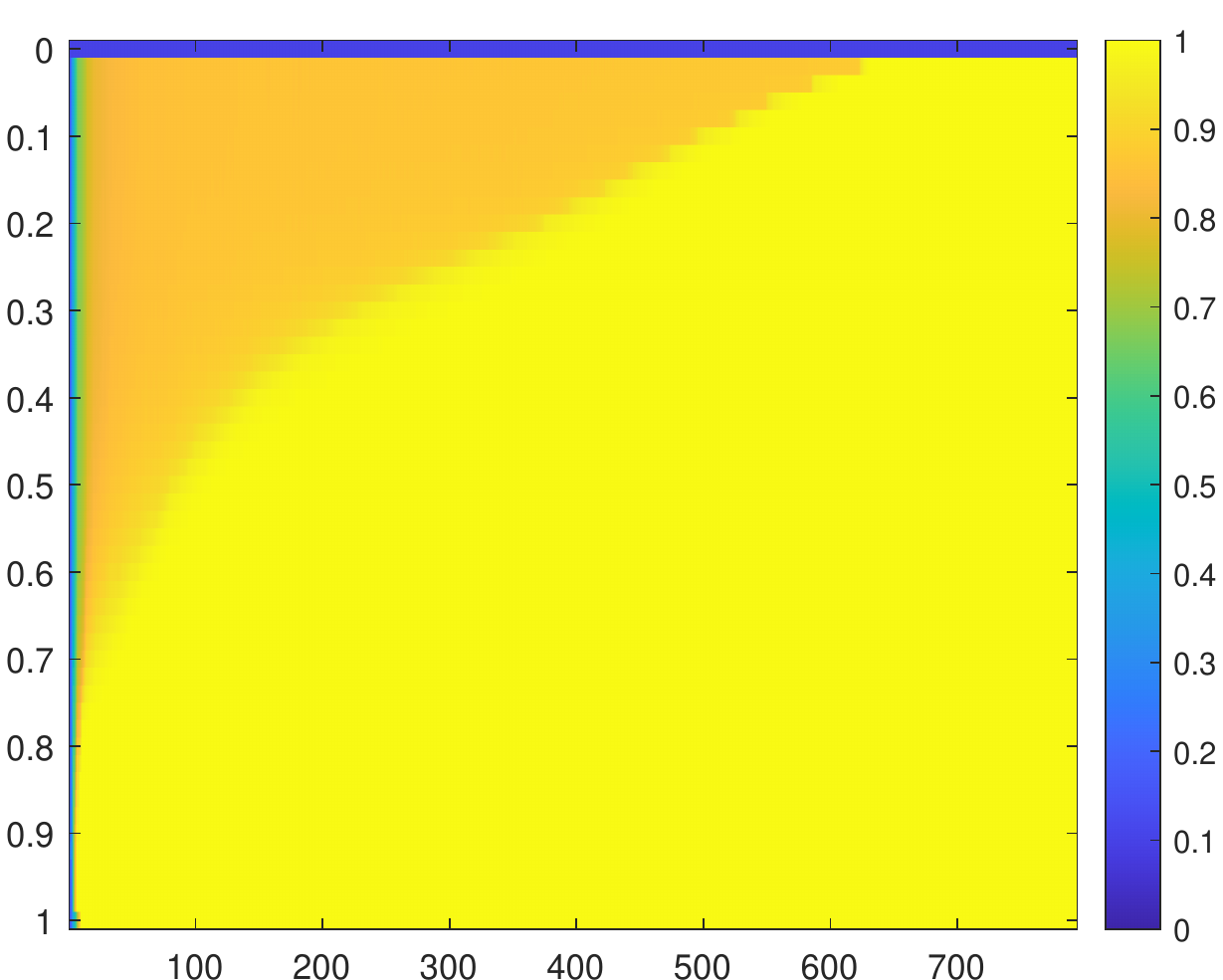}  &
\includegraphics[width=.3\linewidth,trim={4.1cm 10cm 4.4cm 9.7cm},clip]{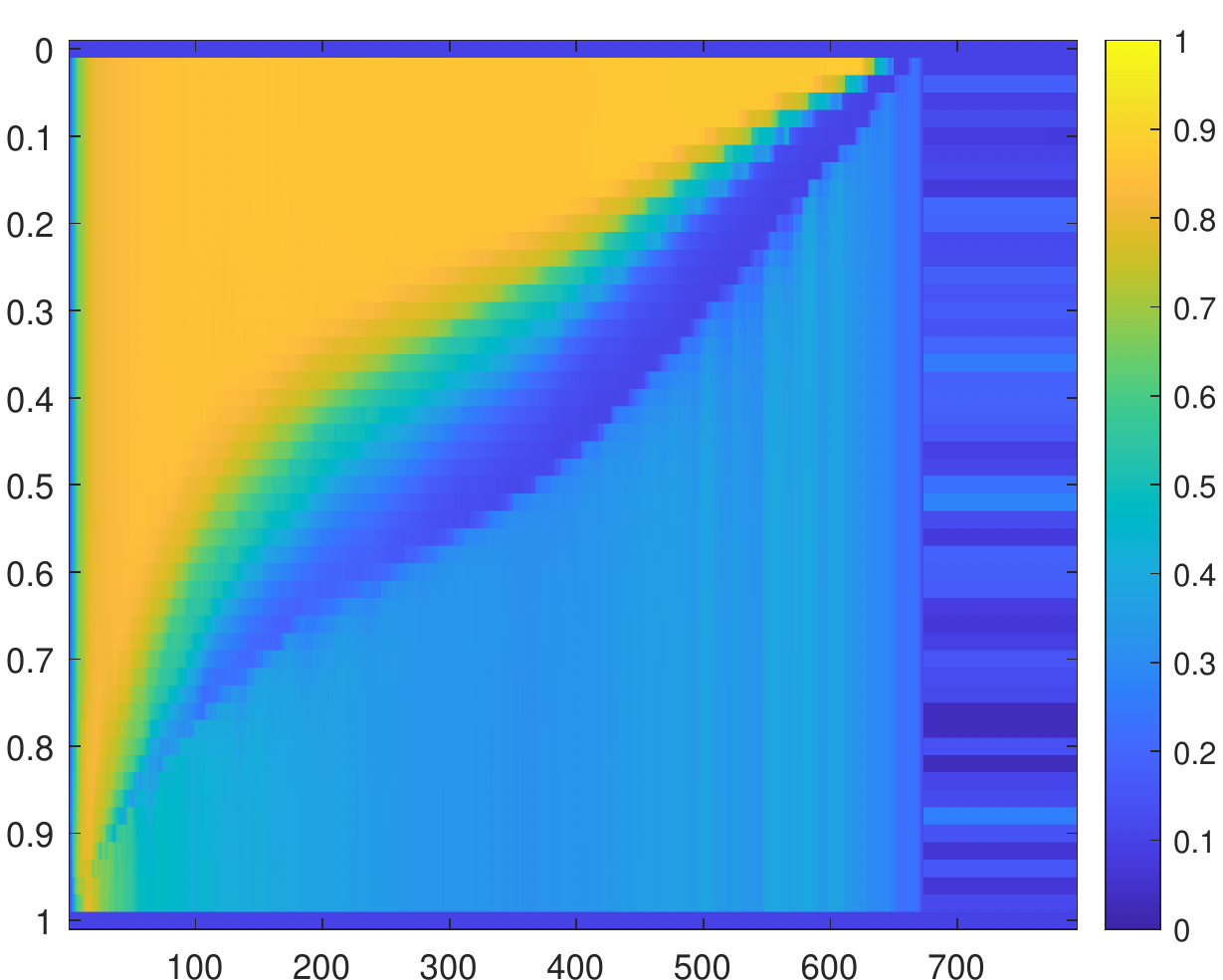} &
\includegraphics[width=.3\linewidth,trim={4.1cm 10cm 4.4cm 9.7cm},clip]{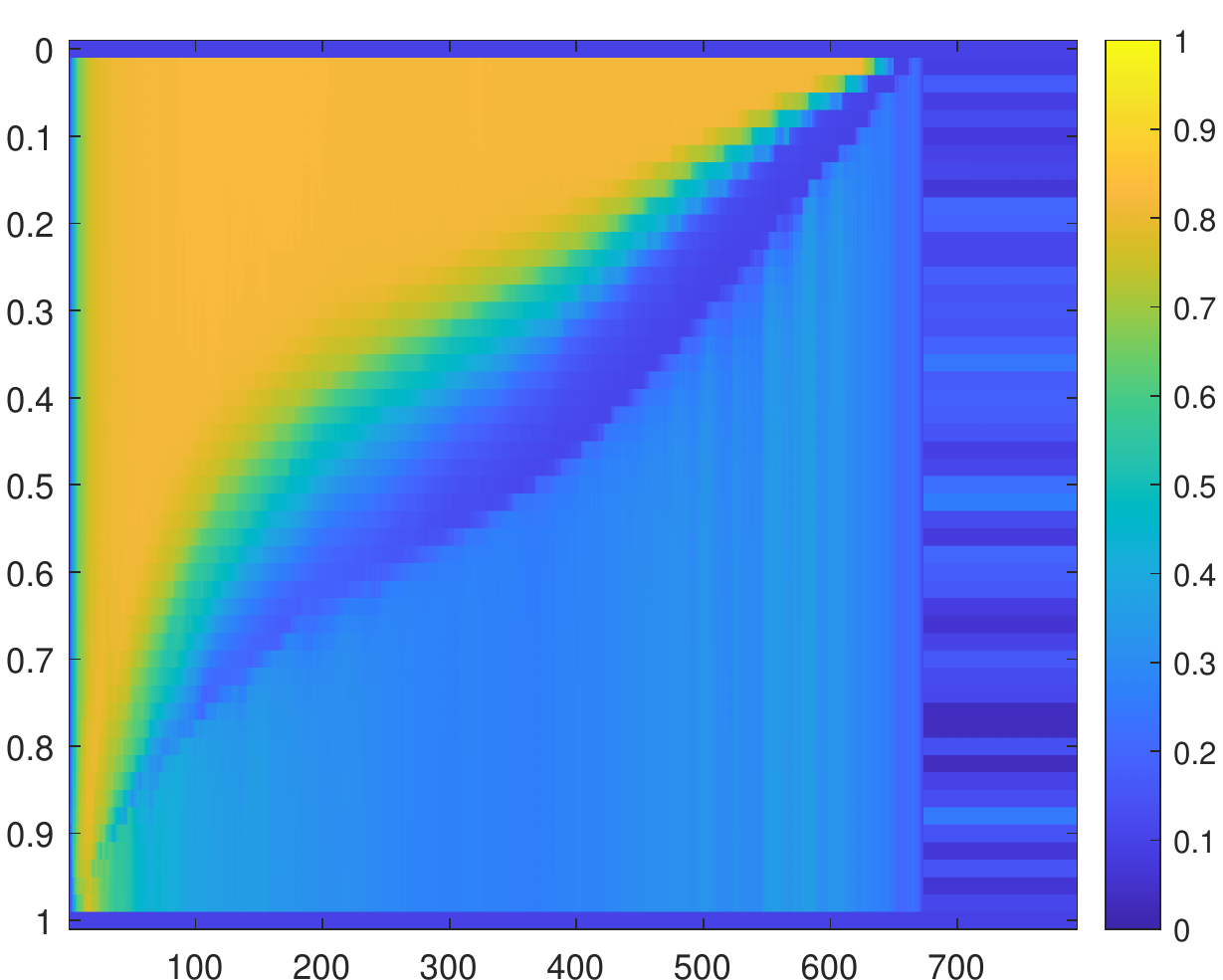} \\
 \hline
\end{tabular}
\caption{Classification accuracy rate (colour coded) over the hyper-parameter space $n_e=[1:1:d_{\mathbf{z}}]$ (abscissa) and $\alpha=[0:0.02:1]$ (y-axis), with comparison to KNN classification results using metric learning \cite{JMLR:v9:vandermaaten08a,10.5555/3045390.3045650,Eusipco2022Collas}. 
}
\label{fig:heatmap}
\end{figure*}

Figure \ref{fig:heatmap} shows  classification accuracies  colour coded as heat maps computed for sets  $\mathcal{I}^{\mathbf{x}}_{\mathbf{y}}$, $\mathcal{I}^{\mathbf{x}}_{\mathbf{0}}$ and $\mathcal{I}^{\mathbf{x'}}_{\mathbf{0}}$, for a range of values $0\leq\alpha\leq 1$ and $n_e\in\lbrace 1,2,\cdots,d_{\mathbf{z}}\rbrace$. 
In practice, the hyper-parameters $n_e$ and $\alpha$ can be chosen with grid search on these heat-maps computed with $\mathcal{I}^{\mathbf{x}}_{\mathbf{0}}$ (acting as validation set) in order to get the best accuracy possible  for the minimum number $n_e$ of principal components that affect the computational cost of our model. 
 As expected, results on set $\mathcal{I}^{\mathbf{x}}_{\mathbf{y}}$ often reach perfect accuracy (Acc=1).   The classification accuracy for sets $\mathcal{I}^{\mathbf{x}}_{\mathbf{0}}$ and  $\mathcal{I}^{\mathbf{x}'}_{\mathbf{0}}$ has a similar behaviour showing that our approach generalised well to unseen features. 

For  MNIST dataset,  heat-maps look smooth on the hyper-parameter space and we see that increasing $\alpha$ allows to concentrate classification efficiency on the first $16$ eigenvectors  for $\mathcal{I}^{\mathbf{x}}_{\mathbf{0}}$ and  $\mathcal{I}^{\mathbf{x}'}_{\mathbf{0}}$  (cf. Fig. \ref{fig:heatmap}). 
The classification accuracy for $\mathcal{I}^{\mathbf{x}}_{\mathbf{0}}$ and  $\mathcal{I}^{\mathbf{x}'}_{\mathbf{0}}$ decreases when too many eigenvectors are used: our model approximates (and converge to) the identity function (when $n_e=d_{\mathbf{z}}$, cf.   Eq. (\ref{eq:decoder}) ) and therefore the model output corresponds to its input which does not provide any class prediction for $\mathcal{I}^{\mathbf{x}}_{\mathbf{0}}$ and  $\mathcal{I}^{\mathbf{x}'}_{\mathbf{0}}$. However, when compressing information in the encoder (i.e. choosing $n_e << d_{\mathbf{z}}$), a class prediction appears as part of the output. For fair benchmark comparison, our models $\mathcal{M}_{0.9}^{16}$ and $\mathcal{M}_{0.02}^{618}$ are retrained\footnote{Demo code available at 
\url{https://github.com/Roznn/DEC}} on the full training set (no data augmentation), and classification accuracy is computed for the full provided test set (cf. Tab \ref{tab:mnist:benchmark}).
\vspace{-0.5cm}
\begin{table}[!h]
\begin{tabular}{lcc}
Model & Accuracy & \# Trainable para. \\
\hline
Ours $\mathcal{M}_{0.9}^{16}$ &0.8093 &12704 \\
Ours $\mathcal{M}_{0.02}^{618}$ & 0.8541 &490692 \\
Efficient-CapsNet \cite{EfficientCapsNet2021} &0.99 &161000 \\
LeNet \cite{MNIST} & 0.99& 60000\\
\hline
\end{tabular}
\caption{Benchmark on MNIST. \footnotesize{Accuracy  result reported for our models $\mathcal{M}_{0.9}^{16}$ and $\mathcal{M}_{0.02}^{618}$ (trained here on the $N=60,000$  images) is computed using the full test set ($N'=10,000$ test images). Its number of trainable parameters corresponds to the number of elements in matrix $\mathrm{U}_{\alpha}^{n_e}$ (i.e. $n_e \times d_{\mathbf{z}}$). While not competing for acuracy  with deep learning models, our results with our models   took  less than 2 seconds for each  $\mathcal{M}_{0.9}^{16}$ and $\mathcal{M}_{0.02}^{618}$ to compute -both training and testing - in this experiment highlighting how frugal our model is. }}
\label{tab:mnist:benchmark}
\end{table}
\vspace{-0.5cm}
\section{Conclusion}

We have introduced a new linear, PCA inspired, classifier  that has encouraging accuracy, can learn from small training datasets as well as larger ones. It has very low computational complexity compared to deep learning models.  
Future work will investigate combining multiple models learning from smaller image patches (as done in CNNs) to improve accuracy on image dataset (ensemble learning), and what alternative strategy to grid search could be used to select  hyper-parameters.

\vfill\pagebreak


\end{document}